\documentclass[10pt,twocolumn,letterpaper]{article}

\usepackage{amsmath,amsfonts,bm}

\def\eqref#1{equation~\ref{#1}}
\def\1{\bm{1}}

\DeclareMathAlphabet{\mathsfit}{\encodingdefault}{\sfdefault}{m}{sl}
\SetMathAlphabet{\mathsfit}{bold}{\encodingdefault}{\sfdefault}{bx}{n}

\usepackage[finalversion]{cvpr}      % To produce the REVIEW version
\definecolor{cvprblue}{rgb}{0.21,0.49,0.74}
\usepackage[pagebackref,breaklinks,colorlinks,allcolors=cvprblue]{hyperref}

\usepackage{graphicx}
\usepackage{tabularx}
\usepackage{booktabs}   % for \toprule, \midrule, \bottomrule
\usepackage{array}      % for better column formatting
\usepackage{url}
\usepackage{diagbox}
\usepackage{multirow}

\newcommand{\enquote}[1] {"#1"}
\newif\ifdebug
\debugfalse % \debugfalse || \debugtrue

\ifdebug
   \usepackage{showframe}
\fi

\newcolumntype{Y}{>{\centering\arraybackslash}X}

\def\paperID{16428} % *** Enter the Paper ID here
\def\confName{CVPR}
\def\confYear{2026}

\title{CRoSS: A Continual Robotic Simulation Suite for Scalable Reinforcement Learning with High Task Diversity and Realistic Physics Simulation}

\author{Yannick Denker \\
Department of Computer Science\\
Fulda University of Applied Sciences\\
Leipziger Str. 123, 36037 Fulda, Germany \\
{\tt\small yannick.denker@cs.hs-fulda.de}
\and
Alexander Gepperth\\
Department of Computer Science\\
Fulda University of Applied Sciences\\
Leipziger Str. 123, 36037 Fulda, Germany \\
{\tt\small alexander.gepperth@cs.hs-fulda.de}
}

\begin{document}
\maketitle

\begin{abstract}
   Continual reinforcement learning (CRL) requires agents to learn from a sequence of tasks without forgetting previously acquired policies.
   In this work, we introduce a novel benchmark suite for CRL based on realistically simulated robots in the Gazebo simulator.
   Our Continual Robotic Simulation Suite (CRoSS) benchmarks rely on two robotic platforms: a two-wheeled differential-drive robot with lidar, camera and bumper sensor, and a robotic arm with seven joints. The former represent an agent in line-following and object-pushing scenarios, where variation of visual and structural parameters yields a large number of distinct tasks, whereas the latter 
   is used in two goal-reaching scenarios with high-level cartesian hand position control (modeled after the Continual World benchmark), and low-level control based on joint angles.
   For the robotic arm benchmarks, we provide additional kinematics-only variants that bypass the need for physical simulation (as long as no sensor readings are required),
   and which can be run two orders of magnitude faster.
   CRoSS is designed to be easily extensible and enables controlled studies of continual reinforcement learning in robotic settings with high physical realism, and in particular allow the use of 
   almost arbitrary simulated sensors.
   To ensure reproducibility and ease of use, we provide a containerized setup (Apptainer) that runs out-of-the-box, and
   report performances of standard RL algorithms, including Deep Q-Networks (DQN) and policy gradient methods. This
   highlights the suitability as a scalable and reproducible benchmark for CRL research.
\end{abstract}

\section{Introduction}
Continual reinforcement learning (CRL) extends the classical reinforcement learning (RL) framework to settings where an agent 
must learn a sequence of tasks over time without access to previous environments.
In contrast to standard RL, where the environment is typically assumed to be stationary and training data can be revisited indefinitely, 
CRL requires agents to continuously integrate new knowledge while preserving performance on previously encountered tasks. 
%Likewise, modern deep learning (DL) models are predominantly effective in static settings that satisfy the i.i.d. assumption, 
%such as image classification \cite{affonso2017deep,li2018deep,li2019deep,wang2021comparative,sharma2022deep} or language modelling \cite{jozefowicz2016exploring, federico1998language}, 
%where training and test distributions remain closely aligned.
This continual setting is relevant for real-world applications such as robotics, 
autonomous driving, or online decision-making systems, where environmental conditions change over time and retraining from scratch is infeasible.

A core difficulty in CRL is catastrophic forgetting \cite{mccloskey1989catastrophic, ratcliff1990connectionist, nguyen2019toward, li2019learn}, 
where newly acquired knowledge interferes with previously learned policies. Although this problem is similar to to the one faced in Continual Learning (CL), 
which deals with learning from non-stationary distributions, but without environment interaction, many established CL methods cannot easily be generalized to CRL.
In particular, the assumption that each task is composed of well-defined sample \textit{classes}, 
and that task onsets and classes are known to the learner, is what distinguishes CL from CRL.
%Addressing forgetting while maintaining plasticity for learning new skills remains a central challenge. 
%Despite numerous algorithmic advances, ranging from replay-based approaches and regularization techniques to generative or rehearsal-free methods, 

\subsection{Motivation}
Current CRL research commonly relies on benchmarks such as robotic reaching suites (e.g., Continual World \cite{wolczyk2021continual}) 
and sequential game-based environments (e.g., Atari-based continual learning setups \cite{abbas2023loss, delfosse2024hackatari}).
While these benchmarks have been valuable for early progress, they exhibit several limitations that constrain their applicability.
Robotic CRL benchmarks typically offer only a small number of distinct tasks, with limited support for sensors, plugins or actuator extensions. 
Many of these environments also do not work "out of the box", requiring substantial installation effort and 
relying on simulation backends that are difficult to configure, distribute, or extend.
For Continual World, observations are low-dimensional, and the tasks do not really require learning in the first place, 
since the robot arm could be controlled directly using inverse kinematics. Additionally, the simulation is kinematics-only, containing no real physics engine.
Game-based continual RL setups, including Atari, provide high-dimensional observations and actions with a wide variety of tasks. 
However, they operate in noise-free, fully deterministic environments that lack physical realism.
Tasks are very diverse, but their intrinsic difficulty is very high, making it difficult to separate intrinsic difficulty from the difficulty of continual learning.
We thus require a benchmark suite that incorporates out-of-the-box readiness, physical realism, a well-supported and extensible simulation backend, and a large number of tasks with high diversity.

Our goal is to close this gap by introducing the Continual Robotic Simulation Suite (\textbf{CRoSS}), a Gazebo-based CRL benchmark suite that combines 
realistic robotic simulation with massive task diversity while providing a reproducible, containerized deployment. 
CRoSS is designed around two complementary robotic platforms: 
First of all, a {two-wheeled differential-drive robot} is used in \textit{multi-task line-following} (MLF) and \textit{multi-task pushing-objects} (MPO) scenarios, where visual and structural parameters (e.g., line shape, textures, colors) are systematically varied to produce hundreds of distinct (but relatively simple) tasks.
On the other hand, a {7-d.o.f. robotic arm} is evaluated in \textit{high-level reaching} (Cartesian end-effector control in six discrete directions mirroring the setup of Continual World) and \textit{low-level reaching}, where the agent directly manipulates the seven arm joints, increasing task dimensionality and realism.

Our benchmark design leverages the Gazebo-Transport middleware communication between sensors, actuators, and agents.
Since Gazebo and the Robot Operating System (ROS) are compatible through a standardized bridge, 
this control interface can be directly applied to physical robots with minimal modification, supporting extensions toward sim-to-real continual learning. 
Our environments can be used as drop-in replacements for standard Gymnasium tasks. 
The environment manager follows the same API conventions as Gymnasium making the benchmark directly compatible with existing RL pipelines and libraries.
Through containerization, installation is simplified across a wide range of platforms.
%By combining controlled variability with physical realism, our benchmark bridges the gap between simple synthetic CRL tasks and realistic robotic learning, 
%while maintaining reproducibility and scalability.

%The benchmark is thus designed to support controlled studies of three central CRL challenges: forgetting (quantifying the degradation of previously learned skills after training on new tasks), 
%transfer (measuring how experience in prior tasks facilitates faster or better learning on subsequent tasks) and scalability (evaluating how algorithms handle growing task amounts in memory requirement and training time).
%By combining controlled variability with physical realism, our benchmark bridges the gap between synthetic CRL testbeds 
%and realistic robotic learning environments while providing full reproducibility.

\subsection{Contribution}
We introduce a new benchmark suite for CRL with the following key contributions:

\par\smallskip\noindent\textbf{Realistic robotic environments:} We present two Gazebo-based robotic platforms, a 2-wheeled differential-drive robot and a 7-d.o.f. robot arm, enabling embodied CRL experiments in physically consistent simulations.
   \par\noindent\textbf{High task diversity and scalability:} By varying visual and structural parameters, our benchmark generates hundreds of distinct tasks, allowing controlled studies of forgetting, transfer, and scalability in continual learning.
   \par\noindent\textbf{Multiple control modalities:} The robotic arm is controlled both in cartesian and joint-space mode, allowing varying action complexities.
   \par\noindent\textbf{Standardized, reproducible and extensible setup:} All environments are provided in a containerized (Apptainer) format that runs out of the box on any Linux system, using a well-supported simulation back-end that can be extended by a variety of community-developed plugins.
   \par\noindent\textbf{Baseline evaluations and metrics:} We benchmark standard CRL algorithms, including DQN and REINFORCE, and provide evaluation metrics for measuring forgetting, transfer, and final performance.
   \par\noindent\textbf{Simulation-to-real compatibility:} The benchmark's communication framework is built on Gazebo-Transport, which is inter-operable with ROS via an existing bridge, making it straightforward to port trained agents or environments from simulation to hardware platforms.

\noindent Together, these contributions establish a scalable, realistic, and easily extensible foundation for advancing continual reinforcement learning research in robotic domains.
All benchmark environments, experiment scripts, and documentation are available in our public repository\footnote{\url{https://github.com/anon-scientist/continual-robotic-simulation-suite/}}.

\section{Related Work}
%Continual reinforcement learning (CRL) extends traditional RL to settings, where agents encounter a sequence of tasks and must learn incrementally 
%without revisiting past experiences.
Comprehensive surveys have outlined the current state and challenges 
of the CRL field \cite{khetarpal2022towards, lyu2019advances, hadsell2020embracing}.
These works summarize existing methods and highlight the open problems in scalability, stability, 
and evaluation that motivate continued research in realistic and dynamic environments.
%Early algorithmic efforts focused on mitigating catastrophic forgetting through regularization-based methods 
%such as weight consolidation or parameter isolation \cite{kirkpatrick2017overcoming, zenke2017continual, schwarz2018progress}, 
%and replay-based approaches that reuse stored or generated samples to retain old knowledge \cite{riemer2018learning, rolnick2019experience, van2019three, gepperth2021gradient}. 
More recently, generative replay and task-agnostic methods have emerged that adapt without explicit task boundaries (e.g.,\cite{lesort2020continual, khetarpal2020options, wolczyk2021zero})
While these advances have improved algorithmic robustness, evaluation often remains confined to synthetic or low-dimensional environments 
that do not reflect the challenges of embodied learning.

Benchmark design has played a critical role in the development of CL algorithms. Classical benchmarks such as Split-MNIST, 
Permuted-MNIST, or CIFAR-100 task sequences \cite{deng2012mnist, xiao2017fashion, kirkpatrick2017overcoming, zenke2017continual, van2019three} 
primarily target supervised continual learning.
In CRL, early environments such as Atari-100k or ProcGen \cite{cobbe2020leveraging, bellemare2013arcade} provided diverse visual tasks, 
though they lack the continuous control and physical realism required for robotic studies.
The Continual World benchmark \cite{wolczyk2021continual} extended the well-known Meta-World suite \cite{yu2020meta} to a continual setting, 
defining sequences of manipulation tasks controlled through simple 3D end-effector movements.
A recent survey of continual RL benchmarks \cite{pan2025survey} provides a broad overview of existing CRL benchmarks.
These can be roughly grouped into game-based environments \cite{delfosse2024hackatari, powers2022cora, tomilin2023coom, kuttler2020nethack, kim2025uncertainty, samvelyan2021minihack, johnson2022l2explorer, nekoei2021continuous, lomonaco2020continual}
which offer visual diversity but lack realism and robotic or physics-based benchmarks \cite{wolczyk2021continual, yang2022evaluations, parisi2020online, isele2018selective, scheel2022urban}, 
which provide physical interaction but typically only few tasks, simplified observations such as perfect 3D positions or lack mechanisms for scalable task generation.
%While Continual World established a valuable foundation for CRL evaluation, it remains limited in task complexity, scalability,-and physical realism, 
%motivating the development of more representative robotic environments.
Frameworks such as MuJoCo, PyBullet, and Gazebo \cite{todorov2012mujoco,coumans2016pybullet, koenig2004design, ferigo2020gym} 
have enabled scalable robotic experiments, though most have been used for isolated single-task RL.

\section{Simulated robots}
\subsection{7-d.o.f. robot arm}\label{sec:arm}
This scenario employs a simulated, seven-degree-of-freedom (7-d.o.f.) robotic arm modeled after the Franka Emika Panda, see \cref{fig:robots} for a visual impression.
The arm consists of seven joints providing a flexible kinematic structure that enables dexterous positioning of the end-effector in three-dimensional space. 
Each joint is actuated independently and reports its angular position and velocity, allowing precise low-level control.
Ground-truth information, such as the end-effector pose, can be retrieved from the simulator for reward generation and policy evaluation.
%
%\noindent Both variants are episodic, with termination triggered upon successful goal reaching or after a maximum number of steps.
%The reward is computed from the Euclidean distance between the current and target end-effector positions.
%This configuration allows studying continual learning under varying control complexity, high-dimensional continuous dynamics, 
%and realistic robot kinematics within a reproducible Gazebo simulation.

\subsection{Two-wheeled mobile robot}\label{sec:robot}
\begin{figure}
\centering
\includegraphics*[page=1,viewport=0cm 0cm 16.9cm 6.5cm,width=0.8\linewidth]{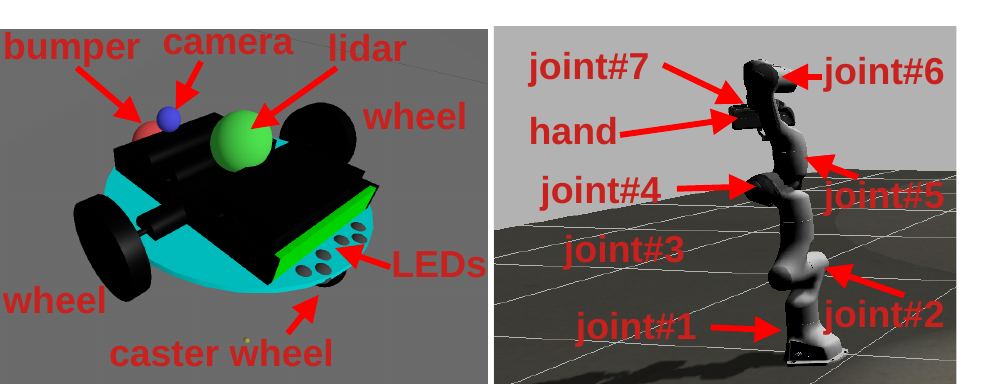}
\caption{\label{fig:robots}
Simulated robots. Left: two-wheeled $3\pi$ robot. Lidar sensors and camera objects are just visual placeholders to indicate their positions. Right: 7-d.o.f. Franka Emika Panda arm. 
}
\vspace{-10pt}
\end{figure}
This is a two-wheeled, differential-drive robot modeled after the Pololu 3Pi, see \cref{fig:robots} for a visual impression. The wheel separation is 10cm, and the wheel radii are 0.5cm each. It is equipped with a contact (bumper) sensor of radius 1cm placed in front, a lidar sensor placed in the center of the robot at a height over the ground of 4cm, and an RGB camera in various configurations. Although the position and orientation can be obtained from the simulator, we only use sensor data that could be obtained from a real robot as well. The lidar sensor covers a forward-looking angle range of $4\pi$ using 500 beams at a frequency of 300 Hz. Camera sensors can be configured to look down or ahead with a variable resolution. In addition to the differential-drive effector, the robot is equipped with 6 LEDs (red, blue, green, magenta, yellow, white) that can be controlled individually.
\section{Benchmarks}
\subsection{Multi-task line following (MLF)}\label{benchmark:mlf}
For this benchmark, the two-wheeled robot (\cref{sec:robot}) is equipped with a line camera that is downwards-oriented for observing the floor ahead of the robot, and has a resolution of 50x3 24-bit color pixels.
The environment is designed to produce a large number (150) of visually distinct tasks through variations in line and ground colors, as well as optional behavior-based modifiers that increase task diversity.
The robot moves on a terrain covered with \textit{tracks} of length 0.75m, consisting of 3 colored lines each (with the center line being slimmer). At the end of each track, there is a wall obstacle.
The line-following tasks are generated from six available colors, resulting in 150 unique tracks. Tracks in which two adjacent lines would have the same color are excluded.
The robot is rewarded for following the left side of the central line, and it equally receives rewards/punishments for turning on the correct lamp at the correct time: in the first half of the track, the robot
must turn on a certain lamp determined by the colors of the three lines, and another for the second half of the track. The robot can measure its distance from the end of the track using its lidar sensor. One task in this benchmark consists of 50 episodes of 30 steps at most, randomly placing the robot at the start of one out of four different tracks (which tracks are used is a property of the task) and letting the robot try to follow the inner line while turning on the correct lamps. From one task to the next, one track is replaced by another that has not been seen before.
The action space is 18-dimensional: the three basic actions "right curve", "left curve" and "straight" are duplicated 6 times for each of the lamps that should be turned on. There is no stopping action. The wheel speeds for all actions are tabulated in \cref{tab:robot-actions}. They are chosen such that forward speed is always constant, and that this constant forward speed is not sufficient to reach the wall in a single episode.
The observation space is represented by a 100x3 RGB image from the robot's line camera. The distance to the end of the track, 
as measured by the lidar sensor, is embedded into the third row of the visual image. For details of reward computation, see \cref{app:reward-mlf}.
This benchmark can be configured in two additional ways besides the \textit{default setting} (DS) described above, leading to progressively easier variants of the benchmark. The goal is to make individual tasks easier, in order to focus more on the difficulty of continual learning with many tasks rather than on the difficulty of learning one particular task.

In the \textit{simplified setting} (SS), the visual signal is transformed to an observation space of 
3x15 mono images encoding the behavior-relevant aspects of wall distance $d_t$, transition pixel position $t_t$, left-line color $L_t$, middle-line color $M_t$ and right-line color $R_t$ in an unambiguous way, see \cref{fig:obs} for details. The information for achieving this is taken from the lidar sensor for $d_t$, from the known properties $L_t$, $M_t$ and $R_t$ of the track the robot is currently placed on, and from image processing of the current observation $\vec o_t$ for the transition pixel position $t_t$.

In the \textit{super-simplified setting} (SSS), the visual signal is transformed as in the SS. In addition, the action space of the DNN controller is reduced to the six actions "LED1" through "LED6". The direction of motion ("straight", "left" or "right", see \cref{tab:robot-actions}) is determined by a non-adaptive controller based on the known value of $t_t$. 
Please see \cref{tab:robot-benchmarks} for an overview of observation and action spaces in all three settings for this benchmark.
\subsection{Multi-task pushing objects (MPO)}\label{benchmark:mpo}
\begin{figure}
\centering
\includegraphics[width=3cm]{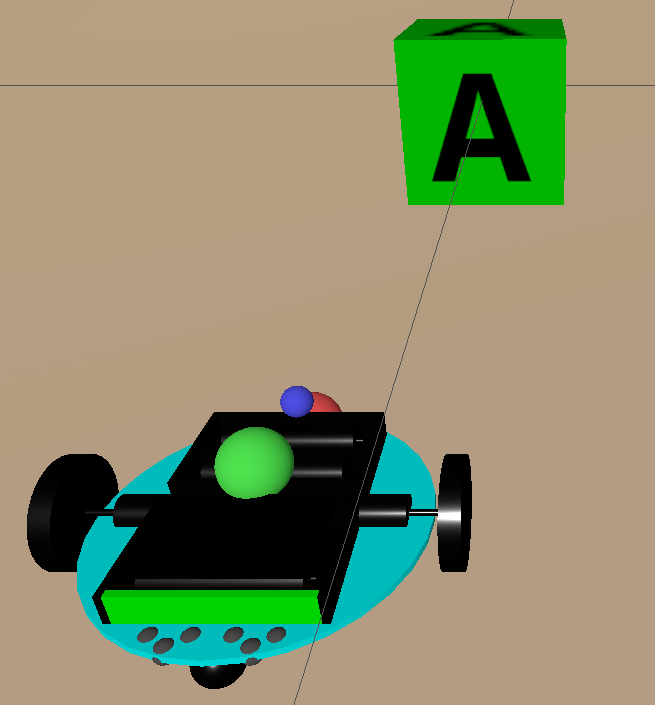}
\includegraphics[width=2.35cm]{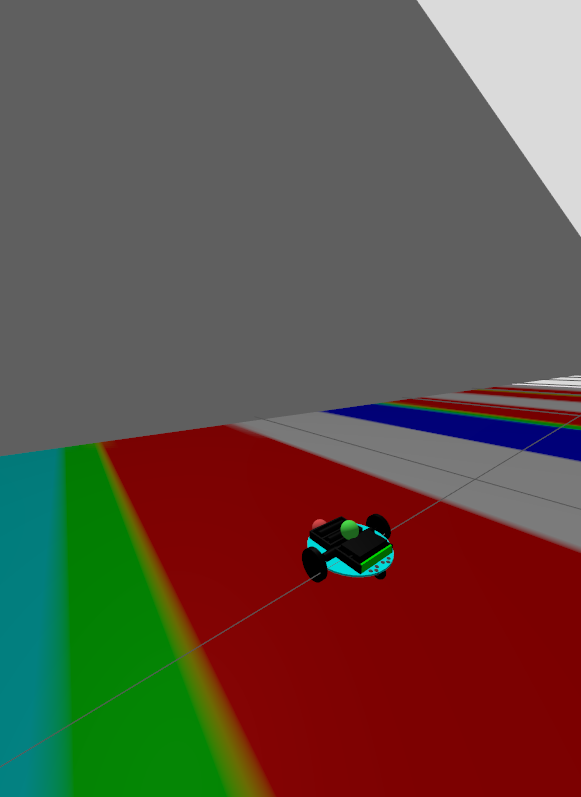}
\caption{\label{fig:benchmarks} Benchmarks for the two-wheeled robot used in this study. Left: multi-task pushing objects (MPO), the robot is approaching geometrical objects with different shapes, colors, and symbols projected onto their faces. Right: multi-task line following (MLF), the robot drives on a ground plane covered with colored lines and limited by a wall. It is supposed to follow the slim centered lines. The robot is the same in both benchmarks but uses a downwards-looking line camera in the MLF benchmark, whereas a forward-looking RGB camera is used in the MPO benchmark.
}
\vspace{-15pt}
\end{figure}
\begin{figure}
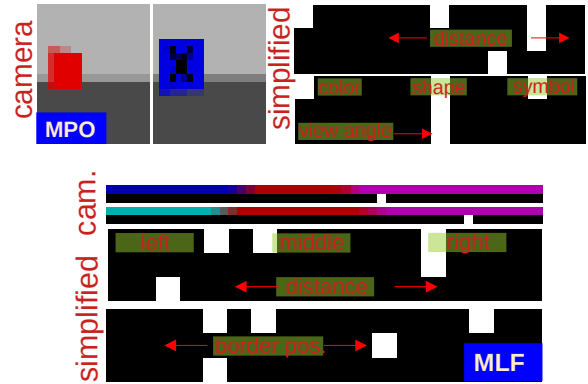

\centering
\includegraphics*[viewport=0cm 0cm 17cm 5cm,width=0.95\linewidth,page=3]{figs/figs.pdf}
\hspace{0.25cm}
\includegraphics*[viewport=0cm 0cm 12cm 6cm,width=0.75\linewidth,page=2]{figs/figs.pdf}
\caption{\label{fig:obs}
Sensor data used for learning in the MLF and MPO benchmarks. Left: two default-setting 20x20 RGB images and two simplified 15x3 mono images from the MPO benchmark. Simplified inputs are population-coded, meaning that the position of the single bright pixel indicates a sensor reading. The uppermost row is split into three population codes for color, shape and symbol of the currently approached object, whereas the next rows encode object distance and viewing angle. Right: Two default-setting 100x2 composite images and two simplified 18x3 images from the MLF benchmark. Composite images integrate line camera data (upper row) and lidar distance to wall (lower row), the latter again population-coded. Simplified images contain three population codes for left/middle/right line color (upper row), distance to wall (middle row) and centering of the left border of the middle line in the image (population-coded). The simplified inputs are helpful because they simplify the problems without sacrificing their massively continual character.
}
\vspace{-22pt}
\end{figure}
In this benchmark, the robot of \cref{sec:robot} is placed before one of 150 possible objects. It is always rewarded for approaching them, and punished or rewarded for driving into them ("pushing"). Whether an object incurs a reward ("pushable object") or punishment ("non-pushable object") if the robot drives into it, depends on the shape, color and symbol place on the faces of the object. One approach episode of at most 30 steps, with a target object of distinct shape, color and symbol, is termed a \textit{track}. Since there are 5 choices for color (white, pink, red, green, blue), 6 choices for symbols ($\#$, A, X, O, plain, H) and 5 choices for shape(sphere, cube, pyramid, diamond, cylinder), there is a total of 150 possible \textit{tracks}. One task consists of four tracks depending on the task, and only one track is replaced by an unseen one from one task to the next, leading to significant task overlap. 
Each episode begins with the robot positioned at a distance of .45m in front of an object, with a randomized orientation of $\pm18^\circ$ relative to the object's direction. This angular offset prevents the robot from reaching the object by chance and requires directed action.
There are four possible actions $a_t$ : "turn left", "turn right", "straight" and "stop", see 
\cref{tab:robot-actions}. Observations $\vec o_t$ are 20x20 RGB images of the robot's front view camera, 
see \cref{fig:obs} for same examples. For details of reward computation, see \cref{app:reward-mpo}.
%
% Fake-inputs+steering, fake-inputs+no-steering, real-input+no-steering
This benchmark can be configured in two additional ways besides the \textit{default setting} (DS) described above, leading to progressively easier variants of the benchmark. The goal is to make individual tasks easier, in order to focus more on the difficulty of continual learning with many tasks.

In the \textit{simplified setting} (SS), the visual signal is transformed to an observation space of 3x15 mono images encoding the behavior-relevant aspects of object distance $d_t$, viewing angle $\phi_t$, color $c_t$, form $f_t$ and depicted symbol $s_t$ in an unambiguous way, see \cref{fig:obs} for details. The information for achieving this is taken from the lidar sensor for $\phi_t$ and $d_t$, as well as the known properties $c_t$, $f_t$ and $s_t$ of the track the robot is currently placed on.

In the \textit{super-simplified setting} (SSS), the visual signal is transformed as in the SS. In addition, the action space of the DNN controller is reduced to the two actions "stop" and "go", where the latter indicates that the full action ("straight", "turn left" or "turn right", see \cref{tab:robot-actions}) should be determined by a non-adaptive controller based on the known value of $\phi_t$. Please see \cref{tab:robot-benchmarks} for an overview of observation and action spaces in all three settings.
\subsection{High-Level Reaching (HLR)} 
This benchmark mirrors the simplified setup used in Continual World and involves reaching target positions with the end-effector (\enquote{hand}) of the simulated Franka Emika Panda arm from \cref{sec:arm}.
Tasks are characterized by the 3D target positions, and episodes always last for 30 steps.
%The challenge lies in precise spatial reasoning, constraint-aware planning, and reward-efficient movement.
The agent controls the robot arm in Cartesian space using 6 discrete actions that move the hand along the three axes ($\pm x$, $\pm y$, $\pm z$).
The observation includes the current and goal 3D positions of the end-effector. Further details about observation and actions spaces are provided in \cref{app:obs_act_HLR_LLR}.
Upon choosing an action, the new target position is passed to PyKDL\footnote{\url{https://docs.ros.org/en/diamondback/api/kdl/html/python/}} 
which computes the corresponding joint angles by inverse kinematics.
These angles are directly applied to the arm using the Gazebo IKPanda plugin. 
If the desired position violates joint limits or is otherwise unreachable, the movement is aborted and a penalty is applied.
For details of reward computation, see \cref{app:reward-high}.
\subsection{Kinematic High-Level Reaching (HLR-K)} \label{sec:kinematic_high_level}
In addition to the physics-based simulation, we provide a kinematic implementation of the High-Level Reaching benchmark that eliminates the need for a simulator entirely.
Instead of executing motion commands on a physical robot model, the system performs direct forward kinematics (FK) computations to update a \textit{"fake"} end-effectors position in cartesian space.
At each step, the agent chooses one of six discrete movement commands, each translating the end-effector in a positive or negative direction along one of the three coordinate axes.
The new position of the end-effector is computed by applying a fixed offset in the chosen axes direction to the previous position.
This results in a lightweight, fully deterministic environment that replicates the kinematic behavior of the simulated arm but omits physical effects such as inertia or collisions.
The reward and termination criteria remain identical to those of the simulated benchmark, allowing direct comparability of results.
Because no physics or message communication needs to be simulated, this version runs orders of magnitude faster (see \cref{app:runtimes} for a comparison), making it ideal for large-scale hyper-parameter searches.
\subsection{Low-Level Reaching (LLR)}
In this novel benchmark, the agent controls the robot arm at the level of joint angles instead of cartesian coordinates.
The goal remains to reach a given 3D target position with the end-effector, requiring coordinated multi-joint motion and precise spatial reasoning.
To this end, an implicit model for local inverse kinematics is learnt around the initial configuration of the arm.
%This setup exposes the full 7 d.o.f. of the arm and requires the agent to learn how to coordinate joint movements to achieve precise positionings of the end-effector.
An episode spans 7 steps, in each of which the agent selects a target angle for a single joint. 
%The selected value is in the range $[0,1]$, which is then linearly mapped to the allowed range of motion for the currently active joint. 
%The mapping is performed using the joint's predefined minimum and maximum angle limits.
The joints are controlled sequentially in a fixed order, where step $s$  controls joint $s$.
The action space remains discrete, with each action representing one out of 5 discrete angles in the joint's allowed angle range.
%The target angle range for this joint is quantized to 5 discrete actions using the joint's predefined minimum and maximum angle limits.
%Ten evenly spaced percentage levels define the available actions, resulting in different effective angular displacements for each joint due to their unique mechanical limits.
%This formulation allows for a compact action representation while preserving the physical constraints of the robot.
The benchmark consists of 8 tasks, each defined by a goal position located at 
varying locations and heights around the arm's workspace.
Observations include the current cartesian position of the end-effector, the cartesian goal position, the joint angles and the index of the joint being controlled.
This additional index helps the agent specialize its output depending on which joint is active at each time step.
Further details about observation and actions spaces are provided in \cref{app:obs_act_HLR_LLR}.

Due to the discrete episode structure and single final reward, we adopt a simplified policy gradient approach similar to REINFORCE, where only the final reward is used for updating the policy.
In our setup, all intermediate rewards are set to $0.0$ and only the final step receives a non-zero reward, with the discount factor $\gamma = 1.0$.
This setup allows for rapid experimentation and emphasizes the learning of accurate motor programs over multiple joints without requiring dense feedback.
For details of reward computation, see \cref{app:reward-low}.
\subsection{Kinematic Low-Level Reaching (LLR-K)}\label{sec:kinematic_low_level}
Similarly to the HLR benchmark, the LLR benchmark includes a kinematic variant based entirely on forward and inverse kinematics (IK) calculations.
Here, the robot arm is represented as a 7-d.o.f. kinematic chain without any physical simulation.
The agent operates directly in joint space, selecting actions that modify joint angles.
The resulting end-effector position is computed using forward kinematics, and the next observation is derived from the new joint and cartesian states.
In contrast to the simulated setup, this version bypasses Gazebo and the Gazebo-Transport messaging layer entirely, 
drastically reducing computation time and hardware requirements.
By isolating the kinematic problem, it allows efficient evaluation of learning dynamics, exploration strategies, 
and policy architectures without the overhead of physics simulation.
%The use of inverse kinematics (IK) primitives also allows for straightforward generation of a large number of distinct tasks.
%By simply varying the initial joint configuration, new reaching problems with different kinematic constraints and trajectories can be created, 
%greatly expanding the task diversity within the same setup.

The kinematic reaching benchmarks therefore complement their simulated counterparts: they preserve the same task definitions, reward functions, 
and termination conditions, but trade physical realism for computational efficiency, making them particularly well suited for rapid experimentation and algorithm development.

\section{Experiments}
Our experiments were conducted on a High-Performance Computing (HPC) cluster consisting of 40 Linux workstations, each equipped with an Nvidia GTX 3090 graphics card. 
%This powerful setup enables the parallel processing capabilities required for large-scale reinforcement learning experiments.
We used TensorFlow 2.14 as the primary deep learning framework to implement and train our reinforcement learning models. 
%In addition, we utilized essential Python libraries such as NumPy for numerical computations. 
All experiments were deployed using Apptainer containers, providing a reproducible environment.
\begin{table}
\centering
\footnotesize
\caption{\label{tab:ind}
Evaluation of randomly picked individual tasks in the MPO and MLF benchmarks for all settings. We observe a small degradation between SSS and DS, which may be explained by increased task difficulty.
MPO results differ between tasks according to whether to target object should be pushed (+10 reward) or not pushed (only small reward for stopping in time).
}
\vspace{-5pt}
\begin{tabular}{|c|cccccccc|}
\hline
task$\rightarrow$       &  51 & 52 & 53 & 54 & 106 & 107 & 108 & 109 \\
\hline
$\downarrow$ setting & \multicolumn{8}{c|}{MLF} \\
\hline
DS                & 1.45 & 1.48 & 1.44 & 1.48 & 1.49 & 1.44 & 1.42 & 1.41  \\
SS                & 1.52 & 1.52 & 1.51 & 1.49 & 1.54 & 1.51 & 1.52 & 1.50  \\
SSS                & 1.61 & 1.58 & 1.61 & 1.61 & 1.57 & 1.62 & 1.61 & 1.61  \\
\hline
$\downarrow$ setting & \multicolumn	{8}{c|}{MPO} \\
\hline
%task$\rightarrow$       &  51 & 52 & 53 & 54 & 106 & 107 & 108 & 109 \\
%\hline
DS                & 21.0 & 11.3 & 22.7 & 12.0 & 21.5 & 11.5 & 21.5 & 12.0  \\
SS                & 21.3 & 11.8 & 20.9 & 12.9 & 22.0 & 11.8 & 21.8 & 11.9  \\
SSS                & 23.5 & 13.2 & 23.9 & 13.9 & 23.0 & 12.5 & 23.6 & 13.7  \\
\hline

\end{tabular}
\vspace{-10pt}
\end{table}

\subsection{Multi-Task Line-Following (MLF)}
Experiments are conducted for each setting: default (DS), simplified (SS) and super-simplified (SSS), see \cref{benchmark:mlf}. The benchmark consists of 150 tasks, which are executed one after the other for 300 (100 for SS and SSS) episodes per task using 30 steps per episode. Epsilon-greedy exploration is used, decreasing $\epsilon$ from either 1.0 (first task) or 0.5 (all other tasks) to a minimal value of 0.2 at the end of a task. A DNN with three hidden layers of size 100 maps observations to actions, where observations depend on the chosen setting, see \cref{tab:robot-benchmarks}). A replay buffer size of 15000 is used for all experiments, roughly 5 tasks' worth of samples. For each task $t'<t$, results from 10 exploitation-only episodes are averaged after training on task $t$.
Due to the large number of tasks, we report evaluation performance on task 1 after training on tasks 1,5,10,50,90, 130 and 150. As an elementary performance measure, we use the average cumulated score per (test) episode, normalized by episode length. All results are obtained by averaging three independent runs with identical parameters. The results are summarized in \cref{tab:3pi-results}.%, and more detailed evaluations are found in \cref{app:eval-mlf}.
%This is the only performance measure, since this benchmark does not lend itself to defining a simple success criterion: simply completing an episode can still imply a very small score due to incorrect LED activation. All results are obtained by averaging three independent runs with identical parameters. The results are summarized in \cref{tab:3pi-results}, and more detailed evaluations are found in \cref{app:a2}.
In addition to the CRL experiments, we also investigated the intrinsic difficulty of individual tasks, randomly training on tasks 1--5 and 105--110 for this purpose. 
In \cref{tab:ind}, we observe that all tasks can be learned to a similar degree of precision.
\begin{table*}[t]
\centering
\footnotesize
\caption{\label{tab:3pi-results}
Experimental results for the MPO and MLF benchmarks. Shown is average reward per step for the MLF benchmark and average cumulated reward for the MPO benchmark, evaluated for task 1 after having trained on task $t$.
The retention of the first task is a simple yet informative measure that retains meaningfulness even when the number of tasks is large. Clear signs of forgetting can be observed for all benchmarks and settings.
}
\vspace{-5pt}
\begin{tabular}{|c|cccccc|cccccc|}
\hline
Benchm. $\rightarrow$ & \multicolumn{6}{c|}{MLF} &  \multicolumn{6}{c|}{MPO} \\
\hline
$\downarrow$ Setting                           & \multicolumn{6}{c|}{evaluated after task} & \multicolumn{6}{c|}{evaluated after task}\\
       &  1 & 5 & 10 & 70 & 120 & 150 & 1 & 5 & 10 & 70 & 120 & 150\\
\hline
DS                & 1.41 & 1.31 & 1.05 & -0.21 & 0.10 & -0.54  & 11.8 & 11.9 & 3.0 & -2.0 & -1.1 & 4.0 \\
SS                & 1.51 & 1.52 & 0.99 & 0.11 & -0.55 & 0.05  & 12.3 & 11.5 & 2.7 & 3.5 & -3.1 & -3.6\\
SSS               & 1.59 & 1.55 & 1.49 & 0.33 & -0.75 & -0.32  & 13.8 & 11.2 & 11.3 & 0.7 & -1.2 & -1.7 \\
\hline
\end{tabular}
\vspace{-10pt}
\end{table*}

\subsection{Multi-task Pushing-Objects (MPO)}
Experiments are conducted for each setting: default (DS), simplified (SS) and super-simplified (SSS), see \cref{benchmark:mpo}. The benchmark consists of 125 tasks (each including four tracks), which are executed one after the other for 300 (100 for SS and SSS) episodes per task using 30 steps per episode. Epsilon-greedy exploration is used, decreasing $\epsilon$ from either 1.0 (first task) or 0.5 (all other tasks) to a minimal value of 0.2 at the end of a task. A DNN with three hidden layers of size 100 maps observations to actions, where observations depend on the chosen setting, see \cref{tab:robot-benchmarks}). A replay buffer size of 15000 is used for all experiments, roughly 5 tasks' worth of samples. For each task $t'<t$, results from 10 exploitation-only episodes are averaged after training on task $t$.
Due to the large number of tasks, we report evaluation performance on task 1 after training on tasks 1,5,10,50, 100 and 125. As an elementary performance measure, we use the average cumulated score per (test) episode. All results are obtained by averaging three independent runs with identical parameters. The results are summarized in \cref{tab:3pi-results}.%, and more detailed evaluations are found in \cref{app:eval-mpo}.
Run-time is approximately 15 minutes per task or approximately or 30 hours in total.

In addition to the CRL experiments, we also investigated the intrinsic difficulty of individual tasks, randomly training on tasks 1--4 and 106--110 for all settings. 
In \cref{tab:ind}, we observe that all tasks where touching the object is expected reach approximately the same reward, and the same holds for tasks where pushing the object is punished, although the reward is lower 
in the latter case because the robot only receives the reward for driving towards the object. We thus conclude that all tasks can be individually solved by the robot, and any performance degradation originates in the sequential nature of learning.

\subsection{High-Level Reaching (HLR)}\label{sec:exp_hlr}
%To evaluate the performance and continual learning behavior of standard RL algorithms on the High-Level Reaching benchmark, 
For the HLR benchmark we conduct a series of experiments using a Deep Q-Network (DQN) as baseline learner.
The overall task configuration is summarized in \cref{tab:hlr_task_setup}, 
comprising ten distinct reaching tasks that differ in their target goal positions within the 3D workspace.

Each experiment consists of a sequential training run across all ten tasks.
For every task $T_i$, the agent is trained for 5000 steps while retaining the learned model parameters between tasks.
After completing each task, the agent is evaluated on all previously encountered tasks ($T_0 . . . T_i$) for 20 test episodes per task to measure retention and transfer effects.
Each full experiment is repeated three times using identical hyperparameters.
The baseline learner is a standard DQN implementation. The DQN agent is implemented in TensorFlow using an Adam optimizer with a learning rate of $1*10^-4$
and a discount factor $\gamma = 0.8$. The network consists of two fully connected layers with 128 and 64 hidden units, respectively.
Training is performed with a batch size of 32, sampling uniformly from a replay buffer whose capacity was varied between 5000, 10000, and 20000 experiences to analyze the effect of memory size.
Exploration follows an $\epsilon$-greedy strategy, starting from an initial $\epsilon = 1.0$ that linearly decayed by $\Delta_\epsilon = 0.0002$ until reaching a minimum of $\epsilon = 0.2$
At the end of each training phase, the current policy is evaluated without further learning updates.

Performance is measured using two complementary metrics: the \textbf{average step reward}, computed as the mean cumulative step reward over 20 evaluation episodes, 
and the \textbf{accuracy}, defined as the fraction of episodes in which the goal is reached within the success tolerance.
After each task, both metrics are computed for all previously seen tasks yielding a continual evaluation matrix that captures both learning and forgetting over time.
The results are aggregated over four independent runs to reduce stochastic variance and are reported in \cref{tab:hlr_results_reward} and more detailed evaluations are found in \cref{app:eval-hlr}.
The final accuracies for all tasks reach approximately \textbf{11\%}, \textbf{23\%} and \textbf{39\%} for buffer sizes of $5000$, $10000$ and $50000$ at the end of training.

To confirm the learnability of individual tasks in isolation, each of the ten tasks is also trained independently (non-sequentially).
Using the same DQN configuration with a replay buffer size of 200, all tasks converge to 100\% accuracy across three runs, 
demonstrating that each task is individually solvable and that performance degradation in the sequential setup is due to catastrophic forgetting rather than task difficulty.
Most parameter fine-tuning and preliminary experiments is conducted in the kinematic variant (see \cref{sec:kinematic_high_level}) of the benchmark.
This approach provides an identical control interface while offering a significant reduction in computational cost,
training runs in the kinematic setup are several orders of magnitude faster than in the full Gazebo simulation (see \cref{app:runtimes} for a comparison).
All final results are subsequently verified in the physical simulation to ensure that the observed learning behavior 
and performance trends remain consistent in realistic, dynamically simulated conditions.

\begin{table*}[t]
    \begin{center}
      \begin{minipage}{0.25\textwidth}
         \caption{Tasks are modeled after those introduced in the \textit{Continual-World} benchmark \cite{wolczyk2021continual}.}
         \vspace{-5pt}
         \label{tab:hlr_task_setup}
         \footnotesize
         \centering
         \begin{tabular}{|l|c|}
             \hline
             Task & Name \\
             &\\\hline
             T1 & hammer \\
             T2 & push wall \\ 
             T3 & faucet close \\ 
             T4 & push back \\ 
             T5 & stick pull \\ 
             T6 & handle press \\ 
             T7 & push ball \\ 
             T8 & shelf place \\ 
             T9 & window close \\ 
             T10 & peg unplug \\
             \hline
         \end{tabular}
      \end{minipage}
      \hfill
      \begin{minipage}{0.70\textwidth}
         \centering
         \caption{Average step reward for the DQN baseline with a replay buffer capacity of 5000 on the kinematic High-Level Reaching (HLR-K) benchmark. The reported values represent averages over multiple evaluation episodes and independent runs with identical parameters}
         \vspace{-5pt}
         \label{tab:hlr_results_reward}
         \footnotesize
         \begin{tabular}{|l|cccccccccc|}
               \hline
               Task & \multicolumn{10}{c|}{Evaluation after}\\
                    & T1       & T2       & T3       &T4        & T5       & T6       & T7       & T8       & T9       & T10   \\\hline
               T1   & 0.71     & 0.70     & 0.70     & 0.70     & 0.70     & 0.72     & 0.64     & 0.61     & 0.62     & 0.58  \\
               T2   & -        & 0.65     & -0.33    & -0.24    & 0.64     & 0.63     & 0.46     & 0.51     & 0.53     & 0.59  \\
               T3   & -        & -        & 0.53     & 0.58     & 0.65     & 0.69     & -1.18    & -0.34    & -1.32    & 0.64 \\
               T4   & -        & -        & -        & 0.77     & 0.77     & 0.77     & 0.44     & 0.42     & 0.47     & 0.43  \\
               T5   & -        & -        & -        & -        & 0.60     & 0.55     & 0.58     & 0.66     & 0.42     & 0.59  \\
               T6   & -        & -        & -        & -        & -        & 0.54     & 0.54     & 0.02     & 0.47     & 0.59  \\
               T7   & -        & -        & -        & -        & -        & -        & 0.65     & 0.65     & 0.65     & 0.65  \\
               T8   & -        & -        & -        & -        & -        & -        & -        & 0.67     & 0.67     & 0.68  \\
               T9   & -        & -        & -        & -        & -        & -        & -        & -        & 0.67     & 0.64  \\
               T10  & -        & -        & -        & -        & -        & -        & -        & -        & -        & 0.54 \\
               \hline
         \end{tabular}
      \end{minipage}
   \end{center}
   \vspace{-20pt}
\end{table*}

\subsection{Low-Level Reaching (LLR)}
%The Low-Level Reaching benchmark evaluates the agent's ability to control the robot arm directly in joint space, 
%requiring coordinated motion across all seven joints.
%In contrast to the high-level Cartesian variant, where a single action moves the end-effector in 3D space, 
%each step in this benchmark corresponds to an individual joint movement.
%A full episode therefore always consists of exactly seven steps, one per joint, after which a new episode begins.
%This structure enforces sequential control over all joints and tests the agent's capacity to learn coordinated multi-joint behavior. 

The experimental procedure is analogous to that of the HLR benchmark, with a few key modifications to accommodate the joint-space formulation.
Each task $T_i$ is trained for 20000 steps, and after completing training on task $T_i$, the agent is evaluated on all previously seen tasks once.
All other hyperparameters, including network architecture, optimizer, learning rate, discount factor, batch size, and exploration parameters, 
remain identical to those used in the HLR benchmark and are described in detail in the that section.

Unlike in the HLR experiment, where the agent receives incremental feedback at every step, the low-level variant employs a REINFORCE-style 
policy gradient approach in which only the final reward at the end of each episode is of relevance.
%This is computed with the euclidean distance and normalized and then reversed so that values near one indicate successful reaching.
Since the true task outcome, the closeness of the end-effector to the goal, can only be determined after all seven joints have been moved, 
it is back-propagated through the entire episode using a discount factor of $\gamma=1.0$, consistent with standard REINFORCE updates for episodic problems.
%This setup encourages the policy to learn coordinated multi-joint sequences that produce an accurate final pose, rather than optimizing for local, step-wise improvements.
Evaluation follows the same general methodology as the HLR benchmark but is adapted to the episodic reward structure.
Performance is reported as the average final reward per step and success accuracy (fraction of episodes reaching the goal within a given tolerance).
Results are averaged over 4 independent runs and summarized in \cref{tab:llr_results_reward}. More detailed evaluations are found in \cref{app:eval-llr}.

Each of the eight low-level tasks is also trained independently using the same DQN configuration and a replay buffer size of 200.
All tasks reach 100\% accuracy across two runs, confirming that they are individually solvable and that performance degradation 
in the sequential setup arises from catastrophic forgetting rather than inherent task difficulty.
Just like HLR, hyper-parameter selection is performed using the kinematic environment variant to enable rapid experimentation, 
and all results are verified in the simulated Gazebo setup.

\begin{table}[t]
    \caption{Final-step rewards for DQN with a replay buffer of 10000 on the Kinematic Low-Level Reaching (LLR-K) benchmark. Reported values are averages over multiple evaluation episodes and independent runs with identical parameters.}
    \label{tab:llr_results_reward}
    \vspace{-15pt}
    \begin{center}
    \footnotesize
        \begin{tabular}{|l|cccccccc|}
            \hline
            Task & \multicolumn{8}{c|}{Evaluation after} \\
                 & T1   & T2    & T3   & T4   & T5   & T6   & T7   & T8 \\\hline
            T1   & 1.0  & 1.0   & 0.22 & -0.05& -0.04& -0.29& -0.01& 0.02 \\
            T2   & -    & 0.55  & 0.59 & 0.23 & 0.18 & 0.19 & 0.50 & 0.24 \\
            T3   & -    & -     & 0.87 & 0.86 & 0.11 & 0.32 & -0.24& 0.66 \\
            T4   & -    & -     & -    & 0.47 & 0.79 & 0.53 & 0.43 & 0.57 \\
            T5   & -    & -     & -    & -    & 1.0  & 1.0  & 1.0  & 0.60 \\
            T6   & -    & -     & -    & -    & -    & 0.52 & 1.0  & 0.52 \\
            T7   & -    & -     & -    & -    & -    & -    & 0.75 & 1.0  \\
            T8   & -    & -     & -    & -    & -    & -    & -    & 0.53 \\
            \hline
        \end{tabular}
    \end{center}
    \vspace{-20pt}
\end{table}

\section{Discussion}
%This section presents and discusses the results obtained from our experiments on the proposed benchmark suite, 
%evaluates the behavior of standard continual learning methods, and highlights the design characteristics that make the suite both challenging and extensible.
\par\noindent\textbf{Principal conclusions} The results of the previous section confirm that our benchmark captures essential aspects of continual reinforcement learning (CRL). 
Across all experiments, standard reinforcement learning methods such as DQN and REINFORCE show strong signs of 
catastrophic forgetting when trained sequentially on the proposed task sequences.
While individual tasks are fully solvable, performance on earlier tasks consistently deteriorates as new ones are introduced, depending on the size of the replay buffer.
On the other hand, replay buffer size cannot be unbounded, and even if it could, a larger replay buffer implies that learning of new policies is slowed down, see also \cite{zhang2017deeper}. 
%This behavior demonstrates that our environments are challenging even for mature algorithms, 
%making them suitable testbeds for investigating new continual learning strategies that explicitly address forgetting, transfer, and stability-plasticity trade-offs.
%Furthermore, the experiments confirm that performance trends are reproducible across both 
%simulated and kinematic variants, underscoring the robustness of the benchmark design.
\par\noindent\textbf{Choice of algorithms} The goal of this article is to describe the advantages of using the benchmark suite, not on particular CRL algorithms. We therefore demonstrate the performance of standard RL algorithms, like DQN and REINFORCE to establish a baseline. Other RL algorithms, like SAC or PPO, are difficult to adapt to the CRL setting since exploration can only be controlled implicitly, as was shown in \cite{icar25}, and are thus not studied here.
\par\noindent\textbf{Improvements over ContinualWorld} Our benchmark extends and generalizes the concepts introduced in Continual World, preserving its simplicity and comparability while addressing several of its limitations.
In contrast to Continual World's simplified control model, where agents operate the robot arm through direct end-effector translations in six directions,
our suite includes both cartesian and joint-space control, providing deeper control hierarchies and more realistic physics.
Additionally, the inclusion of the two-wheeled robot scenarios introduces an entirely new class of embodied tasks beyond manipulation, 
enabling the study of navigation-based continual learning with a rich sensor setup.
Together, these aspects establish CRoSS as a natural step toward more realistic and scalable CRL evaluation.
\par\noindent\textbf{Extensibility and stability} Key advantages of building on Gazebo are extensibility and long-term stability. New robotic setups, sensors, or task environments can be added with minimal effort, 
as Gazebo already provides a wide array of community-supported sensor types, including RGB and depth cameras, lidar and contact sensors, 
that can be directly exposed as observation channels during training. Furthermore, the Gazebo simulator is well-supported by LTS versions for at least the next decade,
offering a stable and standardized platform for further research. 
CRoSS' modular architecture enables straightforward expansion with new robots, tasks, and learning paradigms.
We designed it to be easily integrated into existing Gymnasium-based workflows. 
Our environment manager mirrors the Gymnasium interface (e.g., \texttt{reset()}, \texttt{step()}, termination handling, and observation/action formatting), 
enabling users to train agents with Gymnasium-compatible algorithms without major modification.
Because all scenarios are defined through Gazebo world files and parameterized configuration scripts, additional tasks can be easily implemented.
\par\noindent\textbf{ROS compatibility} Since the benchmark uses the Gazebo-Transport communication system, which is inter-operable with ROS through the official ROS-Gazebo bridge, 
trained policies can easily be transferred to and executed on real-world robots using identical message structures.
This interoperability makes our benchmark not only a simulation platform but also a bridge between simulated and real robotic learning, 
allowing researchers to evaluate sim-to-real transfer in continual reinforcement learning contexts.
\par\noindent\textbf{Bonus: kinematics-only reaching} Both reaching benchmarks include kinematic variants that bypass physical simulation by relying solely on analytical forward and inverse kinematics to compute robot motion.
This simplification preserves the same task structure, control interface, and reward definition while 
eliminating the computational overhead of Gazebo's physical simulation and inter-process communication.
As shown in \cref{app:runtimes}, the kinematic versions achieve a significant speed-up, reducing training time 
by more than an order of magnitude without altering observed learning dynamics.
These variants make large-scale experimentation, hyper-parameter optimization, and algorithmic ablation studies practical 
while still allowing validation in the full physics simulation for realism.
\section{Future work}
The presented benchmark suite could be extended in several ways. First of all, we could define a large number of additional tasks for the reaching benchmarks by replicating existing tasks but varying the initial posture of the arm. Then, the reaching benchmarks could be made more realistic by replacing the end-effector 3D position in the observations by, e.g., data from a camera or lidar sensor. Lastly, a comparison of current state-of-the-art CRL approaches could be conducted using the benchmark suite.

{
    \small
    \bibliographystyle{ieeenat_fullname}
    \bibliography{main}
}

\appendix
\clearpage
\section{Downloading and installing the benchmark suite}

All components of the CRoSS benchmarks are fully open source and available on Github: \url{https://github.com/anon-scientist/continual-robotic-simulation-suite/}.
Each benchmark is distributed as a separate repository, while this main repository serves as the central entry point and documentation hub.

\noindent\textbf{Installation Steps:}

1. Clone the main repository

2. Install or access Gazebo\\
\noindent CRoSS relies on the Gazebo simulator for all physically simulated benchmarks,
Gazebo can be installed or run inside an Apptainer container using the provided definition file.
The container automatically installs Gazebo, Python dependencies, and all required system libraries.

3. Navigate to a specific benchmark\\
\noindent Each benchmark (e.g., High-Level Reaching, Low-Level Reaching, Line Following, Object Pushing) is hosted in a dedicated repository.
Choose your desired benchmark from the list in the main README or project wiki and clone it.

4. Run the benchmark\\
\noindent Follow the installation and usage instructions in the benchmark's own \textit{README.md}.
Each benchmark repository contains detailed configuration options, parameter files, and example scripts for training and evaluation.

\textit{Note: The kinematic variants of the benchmarks do not require Gazebo and can be executed directly in Python without the simulator, offering a faster alternative for algorithm development and hyper-parameter tuning.}

\section{Detailed information about individual benchmarks}
\subsection{Wheel speeds for MLF and MPO}
The wheel speeds for the simulated two-wheeled robot are given in \cref{tab:robot-actions} for the MPO and MLF benchmark.
Please note that the action space for the MLF benchmark is actually 18-dimensional, because each directional control action can be executed in 6 different ways, depending on which LED is turned on.
\begin{table}[ht]
   \centering
   \footnotesize
   \caption{Possible actions and respective wheel speeds for the MPO and MLF benchmarks. For the MLF benchmark, 
   each of the 3 wheel speeds (in m/s) is complemented by one out of 6 LED commands, making the action space 18-dimensional (3 steering actions times 6 LED actions).
   }
   \label{tab:robot-actions}
   \begin{tabular}{|c|ccc|cccc|}
   \hline
    & \multicolumn{3}{c}{MLF benchmark} & \multicolumn{4}{|c|}{MPO benchmark}\\
      \hline
$\downarrow$ wheel                           & straight & left & right & straight & left & right & stop\\
       \hline
      {left}  & 0.2 & 0.15 & 0.25 & 0.4 & 0.3 & 0.5 & 0  \\
      {right} & 0.2 & 0.25 & 0.15 & 0.4 & 0.5 & 0.3 & 0 \\
%      \textbf{LED}         & 1-6 & 1-6  & 1-6  & -   & -   & -   & -\\       
      \hline
   \end{tabular}
\end{table}
\subsection{Observation and action spaces for MLF and MPO}
The observation and action spaces for both benchmarks are strongly impacted by the setting they are used in: DS, SS or SSS, see \cref{tab:robot-benchmarks}. 
In the SS and SSS settings, sensory information (observations) is simplified into a population-coded form that removes the need for dedicated image processing or object detection modules.
In the SSS setting, not only the observation space is reduced but also the action space, since certain aspects of robot control are performed by a non-adaptive controller.
\begin{table}
\caption{\label{tab:robot-benchmarks}
Overview over observation and actions spaces for both benchmarks involving the two-wheeled $3\pi$ robot. The agent actions row indicates the number of distinct actions that can be executed in principle by the robot in a given benchmark. Depending on the chosen setting (DS, SS, SSS), the number of actions controlled by the adaptive controller may be less. For example, in MPO-SSS, the actual heading of the robot is governed by an hard-coded controller, and only the two actions of "stop" and "proceed" are learned.
}
\centering
   \footnotesize
\begin{tabular}{|c|ccc|ccc|}
\hline
benchm.$\rightarrow$           & \multicolumn{3}{c|}{MPO} & \multicolumn{3}{c|}{MLF}\\
\hline
$\downarrow$ property           & DS & SS & SSS & DS & SS & SSS \\
\hline
obs. space &  20x20x3 & 15x3 & 15x3 & 100x3x3 & 15x3 & 15x3 \\
action space &   4      &   4  &   4  &  18     &  18  &   18 \\  
\#(actions)    &   4      &   4  &   2  &  18     &  18  &   6 \\
\hline
\end{tabular}
\end{table}
\subsection{Task structure for MPO}
There is a total of 150 different tracks characterized by the type of obstacle ($O$, choices are 0:pyramid, 1:cylinder, 2:cube, 3:sphere and 4:diamond), the color of the obstacle($C$, choices are 0:white, 1:pink, 2:green, 3:red and 4:blue) as well as the symbol $S$ depicted on the faces of the obstacle (0:hashtag, 1:plain, 2:X, 3:O, 4:H and 5:A). Each track $t$ is thus uniquely characterized by $t = 30\cdot O + 6\cdot C + \cdot S$.
The reward for pushing the obstacle in a given track $t$ is $+10$ for even values of $T$, and $-10$ otherwise. In order to mitigate forgetting somewhat, we construct each task from four different tracks, where only one track ever changes for a new task. Thus, task 0 contains tracks 0,1,2,3, task 1 contains tracks 1,2,3,4, task
2 contains tracks 2,3,4,5 a.s.o. This benchmark can also be configured to contain only one track per task, making the CRL problem even harder.
\subsection{Task structure for MLF}
There is a total of 150 tracks characterized by the colors of the left ($l$), middle ($m$) and right ($r$) line painted onto the ground plane. Each line can have one out of 6 colors (0:red, 1:green, 2:blue, 3:pink, 4:gray, 5:cyan), which gives 150 tasks after omitting infeasible tracks that have two adjacent lines of the same color. Each feasible track $t$ is thus uniquely characterized by $t = 36\cdot l + 6 \cdot m + r, \ \ \ t \in [0,6^3]$. The LEDs to be switched on in the first and second parts of the track were chosen from a uniform random distribution. As in the MPO benchmark, each task contains 4 tracks, and for each new task, only one track is exchanged for another one that has not yet been seen. Like MLF, this benchmark can be configured to contain only one track per task, making it a more difficult benchmark.
\subsection{Observation and action spaces for HLR and LLR}\label{app:obs_act_HLR_LLR}
In the high-level reaching benchmark the agent controls the robot arm at the level of cartesian 3D coordinates for the end-effector. 
The action space consists of six discrete motion commands, each corresponding to a unit movement of the end-effector along one of the principal 
3D axes as seen in \cref{tab:hlr-robot-actions}. Each action moves the end-effector by 0.1 units, corresponding to 10 cm in the simulation.
Observations consist of the current 3D position of the end-effector $(x_{curr},y_{curr},z_{curr})$ along with the goal position $(x_{goal},y_{goal},z_{goal})$.
\begin{table*}[t]
   \centering
   \footnotesize
   \caption{Mapping of discrete actions to their corresponding 3D motion vectors and descriptive labels used in the High-Level Reaching (HLR) benchmark.}
   \label{tab:hlr-robot-actions}
   \begin{tabular}{|l|c|c|c|c|c|c|}
      \hline
      Action           & 1         & 2          & 3          & 4         & 5         & 6 \\\hline
      Direction        & forward   & backward   & left       & right     & up        & down \\
      3D Motion Vector & $(1,0,0)$ & $(-1,0,0)$ & $(-1,0,0)$ & $(1,0,0)$ & $(1,0,0)$ & $(-1,0,0)$ \\
      \hline
   \end{tabular}
\end{table*}

In the low-level reaching benchmark the agent controls the robot arm at joint level.
A configurable parameter determines the number of discrete actions available per joint.
In our main experiments we use five actions, but we also evaluate larger action spaces (e.g., nine actions).
For a given number of actions $A_i$, each joint's full range of motion is uniformly discretized into $i$ possible target angles.
That is, for a joint with minimum angle $\theta_{min}$ and maximum angle $\theta_{max}$, the discrete actions correspond to:
\begin{equation}
   \theta_k = \theta_{min} + k * \frac{\theta_{max} - \theta_{min}}{i-1}, k = 0,1,...,i-1
\end{equation}
The specific joint limits ($\theta_{min}$,$\theta_{max}$) for all seven arm joints are provided in \cref{tab:llr-joint-angles}.
\begin{table*}[t]
   \centering
   \footnotesize
   \caption{Minimum and maximum joint angles (in radians) for all seven joints of the 7-d.o.f. robotic arm modeled after the Franka Emika Panda, used to define the discretized action spaces in the Low-Level Reaching (LLR) benchmark.}
   \label{tab:llr-joint-angles}
   \begin{tabular}{|l|c|c|c|c|c|c|c|}
      \hline
      Joint                          & 1         & 2          & 3          & 4         & 5         & 6        & 7\\\hline
      Minimum angle ($\theta_{min}$) & $-2.897$  & $-1.763$   & $-2.897$   & $-3.072$  & $-2.897$  & $-0.017$ & $-2.897$\\
      Maximum angle ($\theta_{max}$) & $2.897$   & $1.763$    & $2.897$    & $-0.069$  & $2.897$   & $3.752$  & $2.897$\\
      \hline
   \end{tabular}
\end{table*}
Observations include the current and goal end-effector positions $(x_{curr},y_{curr},z_{curr},x_{goal},y_{goal},z_{goal})$, the seven joint angles $(ang_1,ang_2,ang_3,ang_4,ang_5,ang_6,ang_7)$, and an index $j_{curr}$ indicating which joint is currently being controlled, yielding a 14-dimensional observation vector.
\section{Reward computation}
\subsection{Multi-task Line Following}\label{app:reward-mlf}
The reward $r_t$ provides a dense signal based on how well the robot keeps the left edge of the line centered in its camera view, as well as whether the correct LED is turned on.
\begin{align}
   \nonumber
   r_t & = r^{\text{LED}}_t + r^{\text{lf}}_t, \quad r^{\text{LED}}_t = \pm 1 \nonumber \\
   r^{\text{lf}}_t &= 1 - \left|\frac{d_t - \frac{W}{2}}{\frac{W}{2}}\right| \text{, with } d_t \in [0, W] 
   \label{eq:LF_reward_func}
\end{align}
$d_t$ represents the pixel coordinate of the detected left line edge and $W = 100$ the image width. 
$r^{\text{lf}}_t$ peaks when the line edge is centered and decreases linearly toward the image borders.
If the robot loses visual contact with the line (i.e., the left edge is no longer detected), the episode terminates and a penalty of $r^{\text{lf}}_t=-1$ is applied. The reward $r^{\text{LED}}_t$ for turning on the correct LED is 1, and the punishment for turning on the incorrect one is -1.
\subsection{Multi-task Pushing Objects}\label{app:reward-mpo}
A push reward $r_t^{\text p}$ of +10 is given if a pushable object is touched by the contact sensor, and a punishment of -10 in the adverse case. Whenever the contact sensor is active, the episode is terminated.
A small approach reward $r_t^{\text a}$ is given depending on the angle $\phi_t$ in degrees between the robot's forward direction and the direction of the object as detected by lidar. Effectively, this rewards cases where the object is centered in the robot's field of view. If the robot loses sight of the object, expressed by $r_t^\text{a}<0$, a punishment of -1 is given and the episode is terminated. The total reward is computed as
\begin{align}
   r_t &= r_t^{{a}} + r_t^{{p}} \nonumber \\
   r_t^p &= (\pm 10 \text{ if object touched else }0) \nonumber \\ 
   r_t^{{a}}(\vec o_t, a_t) &= (1-|\phi_t/25|)\times r^s_t \nonumber\\
   r_t^s(a_t) &= (0.1 \text{ if } a_t = \text{stop else } 1) \nonumber \\
   \label{eq:BP_reward}
\end{align}
\subsection{Low-level Reaching}\label{app:reward-low}
No intermediate rewards are provided during an episode. Instead, a single reward is given at the final step, after all joint angles have been set. The reward is computed as:
\begin{align}
   r &= 1 - \| \vec p_{final} - \vec p_{goal}\|_2
   \label{eq:RAJ_reward}
\end{align}
where $\vec p_{final}$ is the position of the end-effector after applying all joint angle changes.
This sparse reward structure encourages the agent to learn joint configurations that lead to precise goal-reaching behavior.
\subsection{High-level Reaching}\label{app:reward-high}
The reward is given after each step and is based on the inverse of the Euclidean distance between the current and goal positions:
\begin{align}
   r_t =
   \begin{cases}
      1.0, & \text{if } \vec p_{curr} - \vec p_{goal} < 0.1 \\
      1 - \|\vec p_{curr} - \vec p_{goal}\|_2, & \text{otherwise}
   \end{cases}
   \label{eq:RA_reward}
\end{align}

This results in a maximum reward set to 1 when the robot is at the goal, and decreasing rewards with increasing distance.
An Episode ends when the end-effector reaches the goal (within a margin of $0.1$ units), or a maximum of 30 steps is taken.
Illegal moves (e.g., unreachable positions) result in a small negative reward and no state change.

\section{Runtime reduction through kinematic benchmark variants}\label{app:runtimes}
To quantify the computational benefits of the analytical benchmark variants, we compare the average total experiment completion time for both the physically simulated 
and kinematic versions of the High-Level Reaching (HLR) and Low-Level Reaching (LLR) benchmarks over multiple runs with the same parameter setup.
The results are summarized in \cref{tab:hlr_trainig_times,tab:llr_trainig_times}.

\begin{table}[ht]
   \centering
   \caption{Average total training and evaluation time measured for the high-level reaching benchmarks.}
   \footnotesize
   \label{tab:hlr_trainig_times}
   \begin{tabular}{|c|c|c|}
      \hline
       &  Simulated & Kinematic\\\hline
      \textbf{Average Time} & $\approx$33h 30min& $\approx$ 42min\\\hline
   \end{tabular}
\end{table}

\begin{table}[ht]
   \centering
   \caption{Average total training and evaluation time measured for the low-level reaching benchmarks.}
   \footnotesize
   \label{tab:llr_trainig_times}
   \begin{tabular}{|c|c|c|}
      \hline
                            &  Simulated & Kinematic\\\hline
      \textbf{Average Time} & $\approx$10h 55min& $\approx$ 40min\\\hline
   \end{tabular}
\end{table}

The performance difference arises primarily from the physical simulation overhead inherent to Gazebo.
Even though the simulation was run at 10x real-time speed, faster execution is impractical because the physics simulation becomes a bottleneck.
Each simulation step incurs a non-negligible delay from physics updates, sensor readings, and inter-process communication.

In contrast, the kinematic variants compute each new robot state analytically using forward or inverse kinematics without any physical simulation.
This eliminates physics integration, collision checks, and message passing overhead, allowing instantaneous state updates at many steps per second.
Because the kinematic formulation preserves identical control logic, observations, and reward structure, 
it yields functionally equivalent training behavior while reducing computation time by one to two orders of magnitude.

The physically simulated experiments were conducted once for each parameter configuration to confirm that the kinematic variants 
can serve as accurate substitutes for the full simulations.
As their results were indistinguishable from the kinematic ones, they are omitted.

\section{Extended experimental results}
\subsection{Multi-task Line Following}\label{app:eval-mlf}
This chapter is reserved for extended experimental results and analyses referenced in the main text.
These materials are part of ongoing work and will be included in a future revision.
\subsection{Multi-task Pushing Objects}\label{app:eval-mpo}
This chapter is reserved for extended experimental results and analyses referenced in the main text.
These materials are part of ongoing work and will be included in a future revision.
\subsection{High-level Reaching}\label{app:eval-hlr}
To complement the sequential training results presented in this paper, we also evaluated each High-Level Reaching (HLR) task independently 
to establish the upper performance bound achievable under isolated conditions.
In this setup, the agent is trained on a single task only, without exposure to the others, 
allowing us to determine the best attainable reward in the absence of interference or forgetting effects.
All experiments used the same DQN configuration as in the sequential setup but with a smaller replay buffer of 200 transitions.
Even under this limited memory capacity, the agent successfully learned all individual reaching tasks to completion, consistently achieving near-optimal behavior.
The resulting average step rewards per episode are summarized in \cref{tab:single_task_rewards} on the left.
Because all tasks reached 100\% success, a separate accuracy table is omitted.
The average step reward values shown in the table therefore represent the maximum achievable performance 
for each task and serve as a reference point for interpreting the degradation observed in sequential training.

\begin{table}[ht]
   \caption{Average step rewards per episode for individually trained tasks in the High-Level Reaching (HLR) benchmark (left) and the final reward of each episode in the Low-Level Reaching (LLR) benchmark with an action space of five (center) and nine (right) discrete actions. 
            Each task was trained independently using the same DQN configuration with a replay buffer capacity of 200. 
            All tasks reached 100\% success for the HLR and LLR with 5 actions, establishing the upper reward bound for comparison with sequential training results.
            The success rate for the LLR benchmark with nine discrete actions varies considerably and strongly depends on how fortunate the exploration process was during training.
            The reward is 1 for all low-level reaching tasks where the goal within the specified threshold of $0.1$ units was reached.}
   \footnotesize
   \label{tab:single_task_rewards}
   \begin{center}
      \begin{minipage}{0.15\textwidth}
         \begin{tabular}{|l|c|}
            \hline
            \multicolumn{2}{|c|}{High-Level}\\\hline
            Task & Reward \\\hline
            T1   &  0.70     \\
            T2   &  0.61     \\
            T3   &  0.87     \\
            T4   &  0.77     \\
            T5   &  0.54     \\
            T6   &  0.59     \\
            T7   &  0.58     \\
            T8   &  0.62     \\
            T9   &  0.60     \\
            T10  &  0.59     \\
            \hline
         \end{tabular}
      \end{minipage}
      \hfill
      \begin{minipage}{0.15\textwidth}
         \begin{tabular}{|l|c|}
            \hline
             \multicolumn{2}{|c|}{Low-Level}\\
             \multicolumn{2}{|c|}{5 Actions}\\\hline
            Task & Reward \\\hline
            T1   &  1     \\
            T2   &  1     \\
            T3   &  1     \\
            T4   &  1     \\
            T5   &  1     \\
            T6   &  1     \\
            T7   &  1     \\
            T8   &  1     \\
            \hline
         \end{tabular}
      \end{minipage}
      \hfill
      \begin{minipage}{0.15\textwidth}
         \begin{tabular}{|l|c|}
            \hline
             \multicolumn{2}{|c|}{Low-Level}\\
             \multicolumn{2}{|c|}{9 Actions}\\\hline
            Task & Reward \\\hline
            T1   &  0.59  \\
            T2   &  1     \\
            T3   &  0.70  \\
            T4   &  0.64  \\
            T5   &  0.62  \\
            T6   &  0.58  \\
            T7   &  0.71  \\
            T8   &  1     \\
            \hline
         \end{tabular}
      \end{minipage}
   \end{center}
\end{table}

\begin{figure*}
   \centering
   \includegraphics[width=0.8\textwidth]{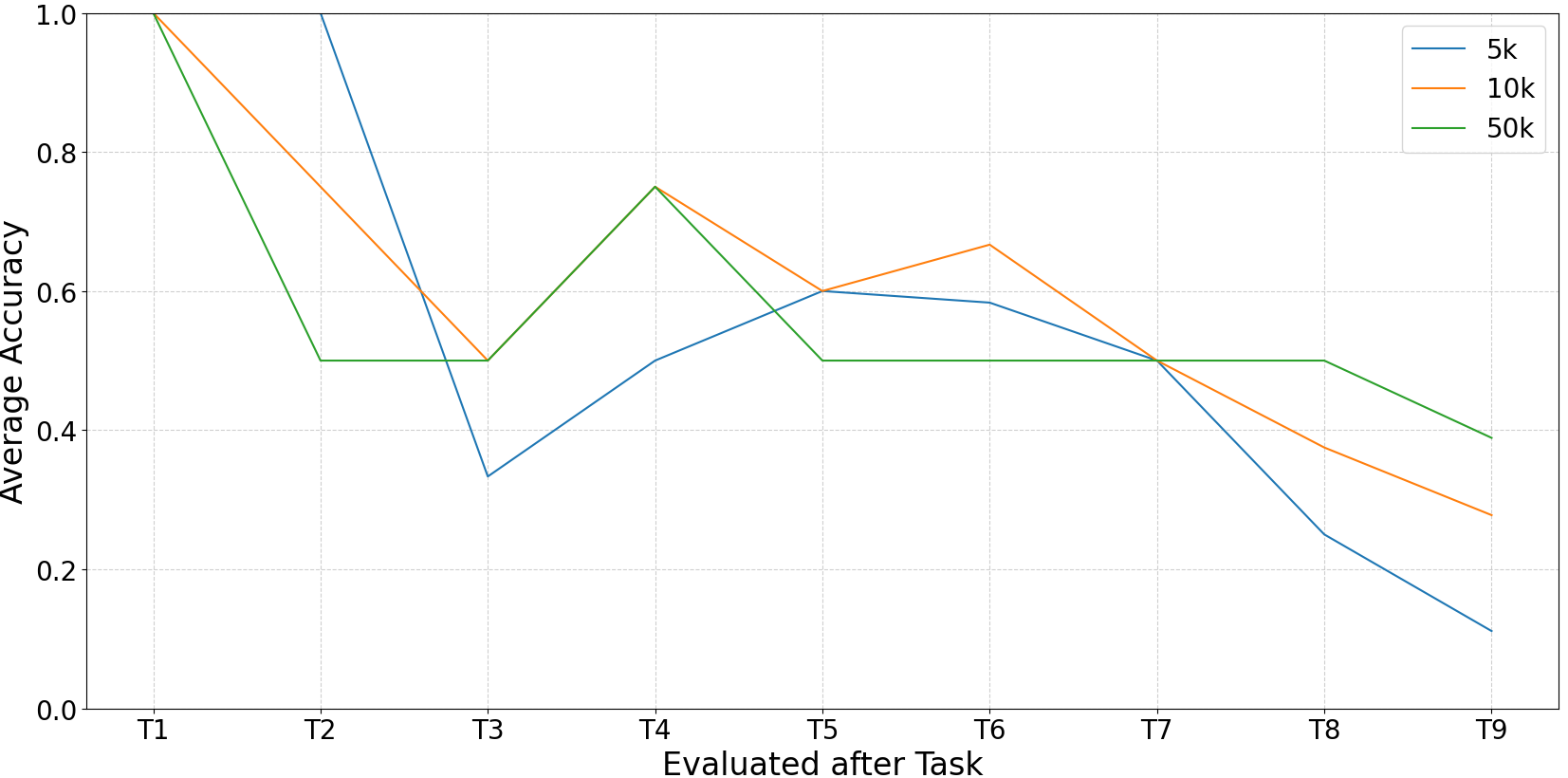}
   \caption{\label{fig:hlr_accuracy_plot}
   Average task accuracy across sequentially introduced tasks in the HLR benchmark with three different buffer sizes.
   As new tasks are added during training, performance on earlier tasks declines, illustrating the effect of catastrophic forgetting.
   }
\end{figure*}

To further analyze the effects of continual training on performance retention, 
we conducted an extended evaluation of the kinematic High-Level Reaching (HLR-K) benchmark across varying replay buffer capacities.
Figure \ref{fig:hlr_accuracy_plot} illustrates the average success accuracy after each sequentially 
learned task for three buffer sizes: 5000, 10000, and 20000.
As shown in the figure, accuracy systematically decreases as additional tasks are introduced.
While the DQN agent initially achieves near-perfect success on early tasks, its performance on previously learned tasks 
drops substantially once training progresses beyond the third or fourth task (see \cref{tab:hlr_results_accuracy}).
This effect is most pronounced for the smallest replay buffer (5000), where the limited memory capacity accelerates forgetting of older experiences.
Larger buffers (10000 and 20000) help retain performance slightly longer, but the overall downward trend persists, 
confirming that even increased replay capacity cannot fully prevent catastrophic forgetting in this setting.

The results highlight the challenging nature of the benchmark, as all tasks are individually solvable in isolation, 
yet sequential training leads to progressive accuracy degradation.
This makes the HLR benchmark particularly well suited for testing methods that aim to improve long-term retention, 
such as generative replay, elastic weight consolidation, or experience balancing strategies.

\begin{table}[ht]
   \centering
   \caption{Task accuracies for the DQN baseline on the kinematic High-Level reaching (HLR-K) benchmark. Reported values represent averages over four runs with identical parameters.}
   \footnotesize
   \label{tab:hlr_results_accuracy}
   \begin{tabular}{|l|llllllllll|}
      \hline
      Task & \multicolumn{10}{c|}{Evaluation after}\\
           & T1       & T2       & T3       & T4       & T5       & T6       & T7       & T8       & T9       & T10   \\\hline
      T1   & 1        & 1        & 1        & 1        & 1        & 0.75     & 0.25     & 0.25     & 0        & 0     \\
      T2   & -        & 0.75     & 0.5      & 0.5      & 1        & 0.5      & 0.5      & 0.25     & 0        & 0.25  \\
      T3   & -        & -        & 0        & 0        & 0        & 0        & 0        & 0        & 0        & 0.5   \\
      T4   & -        & -        & -        & 1        & 1        & 1        & 0.5      & 0.5      & 0        & 0     \\
      T5   & -        & -        & -        & -        & 0        & 0.5      & 0.5      & 0        & 0.25     & 0.25  \\
      T6   & -        & -        & -        & -        & -        & 0.75     & 0.75     & 0.25     & 0        & 0.25  \\
      T7   & -        & -        & -        & -        & -        & -        & 1        & 1        & 1        & 1     \\
      T8   & -        & -        & -        & -        & -        & -        & -        & 0        & 0        & 0     \\
      T9   & -        & -        & -        & -        & -        & -        & -        & -        & 0.25     & 0     \\
      T10  & -        & -        & -        & -        & -        & -        & -        & -        & -        & 1     \\
      \hline
   \end{tabular}
\end{table}

In addition to accuracy, we also analyzed the average step reward obtained during each evaluation on all previously learned tasks.
As described in \cref{app:reward-high}, the reward is normalized between 0 and 1 and represents the relative distance from the end-effector to the goal position, specifically, 
the normalized reduction in Euclidean distance from the start to the goal within an episode. \cref{tab:hlr_results_reward,tab:hlr_results_reward_10000,tab:hlr_results_reward_50000}
show the resulting step rewards over the course of continual training for different replay buffer sizes.

Interestingly, while success accuracy declines sharply as more tasks are introduced, the average step reward often remains relatively high 
and can even increase in later stages of training. This seemingly contradictory behavior arises from two factors inherent to the task design and reward formulation:

\begin{enumerate}
   \item \textbf{Variable episode lengths:} Episodes terminate immediately upon successful goal reaching. As accuracy drops and the agent fails to reach the goal, episodes tend to run for their full length, producing more reward samples per episode. In contrast, successful episodes are shorter, resulting in fewer (but higher-quality) steps, which lowers the per-step average once success is achieved early.
   \item \textbf{Reinforced approach behavior:} The learned policy consistently moves the arm forward toward the goal direction, since all target positions are located in front of the robot's initial pose. Even when the agent fails to complete the final reach, it still performs meaningful approach movements that yield relatively high distance-based rewards. Consequently, the agent can accumulate high per-step rewards without ever registering a \textit{success} event.
\end{enumerate}

Together, these effects explain why average step reward does not directly correlate with accuracy.
An agent may achieve a high normalized reward simply by approaching the target efficiently and lingering 
near the goal without crossing the success threshold, while early successes shorten episodes and reduce the per-step average.
This behavior highlights the importance of interpreting accuracy and reward jointly when evaluating continual reinforcement learning systems, 
especially in environments with distance-based, normalized reward functions.

\begin{table*}[t]
   \centering
   \caption{Average step reward for the DQN baseline with a replay buffer capacity of 10000 on the kinematic High-Level Reaching (HLR-K) benchmark. 
   Reported values represent averages over multiple runs with identical parameters.}
   \footnotesize
   \label{tab:hlr_results_reward_10000}
   \begin{tabular}{|l|llllllllll|}
      \hline
      Task & \multicolumn{10}{c|}{Evaluation after}\\
           & T1       & T2       & T3       & T4       & T5       & T6       & T7       & T8       & T9       & T10   \\\hline
      T1   & 0.70     & 0.71     & 0.71     & 0.70     & 0.71     & 0.70     & 0.72     & 0.72     & 0.63     & 0.63  \\
      T2   & -        & 0.65     & 0.67     & 0.65     & 0.65     & 0.63     & 0.64     & 0.69     & 0.60     & 0.62  \\
      T3   & -        & -        & 0.60     & 0.58     & 0.63     & 0.64     & 0.64     & -0.35    & -1.32    & 0.59 \\
      T4   & -        & -        & -        & 0.77     & 0.77     & 0.77     & 0.76     & 0.76     & 0.60     & 0.60  \\
      T5   & -        & -        & -        & -        & 0.52     & 0.53     & 0.53     & 0.69     & 0.60     & 0.55  \\
      T6   & -        & -        & -        & -        & -        & 0.57     & 0.60     & 0.67     & 0.64     & 0.40  \\
      T7   & -        & -        & -        & -        & -        & -        & 0.65     & 0.65     & 0.65     & 0.65 \\
      T8   & -        & -        & -        & -        & -        & -        & -        & 0.61     & 0.58     & 0.58 \\
      T9   & -        & -        & -        & -        & -        & -        & -        & -        & 0.58     & 0.61 \\
      T10  & -        & -        & -        & -        & -        & -        & -        & -        & -        & 0.57 \\
      \hline
   \end{tabular}
\end{table*}
\begin{table*}[t]
   \centering
   \caption{Average step reward for the DQN baseline with a replay buffer capacity of 50000 on the kinematic High-Level Reaching (HLR-K) benchmark. 
   Reported values represent averages over multiple runs with identical parameters.}
   \footnotesize
   \label{tab:hlr_results_reward_50000}
   \begin{tabular}{|l|llllllllll|}
      \hline
      Task & \multicolumn{10}{c|}{Evaluation after}\\
           & T1       & T2       & T3       &T4        & T5       & T6       & T7       & T8       & T9       & T10   \\\hline
      T1   & 0.71     & 0.71     & 0.70     & 0.71     & 0.70     & 0.70     & 0.71     & 0.70     & 0.71     & 0.73 \\
      T2   & -        & 0.67     & 0.58     & 0.61     & 0.68     & 0.64     & 0.66     & 0.64     & 0.65     & 0.65 \\
      T3   & -        & -        & 0.56     & 0.62     & 0.60     & 0.62     & 0.64     & 0.65     & 0.65     & 0.65 \\
      T4   & -        & -        & -        & 0.77     & 0.77     & 0.77     & 0.76     & 0.77     & 0.76     & 0.76 \\
      T5   & -        & -        & -        & -        & 0.49     & 0.56     & 0.53     & 0.67     & 0.55     & 0.55 \\
      T6   & -        & -        & -        & -        & -        & 0.65     & 0.60     & 0.67     & 0.58     & 0.58 \\
      T7   & -        & -        & -        & -        & -        & -        & 0.68     & 0.66     & 0.65     & 0.65 \\
      T8   & -        & -        & -        & -        & -        & -        & -        & 0.50     & 0.54     & 0.62 \\
      T9   & -        & -        & -        & -        & -        & -        & -        & -        & 0.56     & 0.61 \\
      T10  & -        & -        & -        & -        & -        & -        & -        & -        & -        & 0.53 \\
      \hline
   \end{tabular}
\end{table*}

\subsection{Low-level Reaching}\label{app:eval-llr}
To verify that all Low-Level Reaching (LLR) tasks are individually solvable and to establish the upper reward limit, 
we conducted the same non-sequential training procedure as described for the High-Level Reaching (HLR) benchmark in \cref{app:eval-hlr}.
Each of the eight LLR tasks was trained independently using the same REINFORCE setup with a replay buffer size of 200 
and the resulting average step rewards per episode were included in \cref{tab:single_task_rewards} in the center.
All tasks reached 100\% success, confirming that each task is individually learnable and that performance degradation 
observed in sequential training is solely caused by catastrophic forgetting rather than inherent task difficulty.

\begin{figure*}
   \centering
   \includegraphics[width=0.8\textwidth]{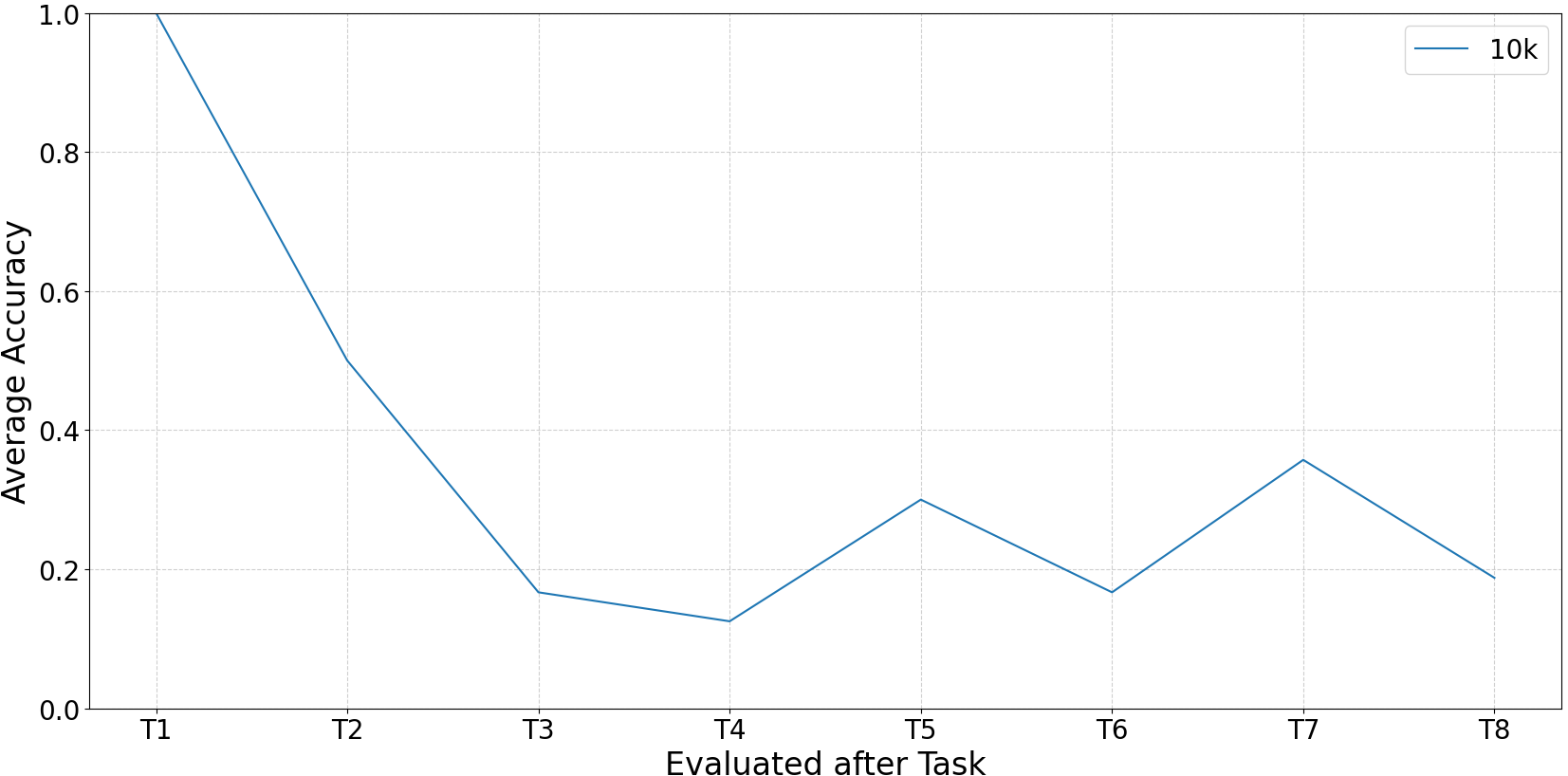}
   \caption{\label{fig:llr_accuracy_plot}
      Average task accuracy across sequentially introduced tasks in the low-level reaching benchmark with a buffer size of 10000.
      As new tasks are added during training, performance on earlier tasks declines, illustrating the effect of catastrophic forgetting.
   }
\end{figure*}

As in the HLR benchmark, we evaluated continual learning behavior by sequentially training the agent on all eight tasks using a replay buffer size of 10000.
\cref{fig:llr_accuracy_plot} shows the average success accuracy after each task. The results mirror the trends observed in the HLR experiments: 
the agent performs well on early tasks but exhibits a progressive decline in accuracy as more tasks are added.

To explore how the granularity of the discrete action space affects learnability, 
we conducted additional experiments varying the number of discrete angular positions available per joint.
The default configuration divides each joint's range of motion into five discrete angles, resulting in a well-balanced action space 
that allows efficient exploration and successful task learning within the given training duration.

When increasing the number of discrete angles to nine, the agent was unable to learn most of the tasks within the same training time.
The resulting average episode rewards (final-step rewards) for this configuration are reported in \cref{tab:single_task_rewards} (left), 
where only two of the eight tasks achieved successful reaching behavior during evaluation. 
This outcome varies considerably between runs and is highly dependent on how effective the exploration process was during training.

With seven joints and five discrete angle values each, the total number of unique configurations is $5^7 = 78125$, 
which remains sufficiently small to explore within the available training time.
However, increasing to nine angles per joint raises this number to $9^7 \approx 4.8 * 10^6$, exceeding what the agent can feasibly explore 
and generalize from within the given training budget. As the action space grows, the probability of revisiting useful configurations decreases dramatically.
While higher angular precision theoretically increases control fidelity, it simultaneously expands the state-action space beyond what can be effectively explored 
using simple exploration methods within realistic training constraints.

\end{document}